\documentclass[runningheads]{llncs}

\usepackage{eccv}
\usepackage{eccvabbrv}

\usepackage{graphicx}
\usepackage{booktabs}
\usepackage[accsupp]{axessibility}

\usepackage{xcolor}
\usepackage{colortbl}
\definecolor{citecolor}{RGB}{34,139,34}
\usepackage[pagebackref=true,breaklinks=true,colorlinks,citecolor=citecolor,bookmarks=false]{hyperref}

\usepackage{orcidlink}

\usepackage{float}
\usepackage{multirow}
\usepackage{makecell}
\usepackage{xfrac}
\usepackage{pifont}
\usepackage{tabularx}

\usepackage[dvipsnames]{xcolor}
\usepackage{colortbl}

\definecolor{dg}{RGB}{0,180,0}
\definecolor{demphcolor}{gray}{.5}
\newcommand{\demph}[1]{\textcolor{demphcolor}{#1}}

\usepackage{amsmath,amsfonts,bm}

\def\eqref#1{equation~\ref{#1}}

\def\1{\bm{1}}

\def\vn{{\bm{n}}}

\def\vt{{\bm{t}}}

\def\vx{{\bm{x}}}

\def\mA{{\bm{A}}}

\def\mI{{\bm{I}}}

\def\mW{{\bm{W}}}
\def\mX{{\bm{X}}}

\DeclareMathAlphabet{\mathsfit}{\encodingdefault}{\sfdefault}{m}{sl}
\SetMathAlphabet{\mathsfit}{bold}{\encodingdefault}{\sfdefault}{bx}{n}

\def\sR{{\mathbb{R}}}

\begin{document}

\title{SCLIP: Rethinking Self-Attention for Dense Vision-Language Inference} 
\titlerunning{Rethinking Self-Attention for Dense Vision-Language Inference}

\author{Feng Wang \and Jieru Mei \and Alan Yuille}

\authorrunning{F. Wang et al.}

\institute{Johns Hopkins University}

\maketitle

\begin{abstract}
Recent advances in contrastive language-image pretraining (CLIP) have demonstrated strong capabilities in zero-shot classification by aligning visual and textual features at an image level.
However, in dense prediction tasks, CLIP often struggles to localize visual features within an image and fails to attain favorable pixel-level segmentation results.
In this work, we investigate in CLIP's spatial reasoning mechanism and identify that its failure of dense prediction is caused by a \textit{location misalignment} issue in the self-attention process.
Based on this observation, we propose a training-free adaptation approach for CLIP's semantic segmentation, which only introduces a very simple modification to CLIP but can effectively address the issue of \textit{location misalignment}.
Specifically, we reform the self-attention mechanism with leveraging query-to-query and key-to-key similarity to determine attention scores.
Remarkably, this minimal modification to CLIP significantly enhances its capability in dense prediction, improving the original CLIP's 14.1\% average zero-shot mIoU over eight semantic segmentation benchmarks to 38.2\%, and outperforming the existing SoTA's 33.9\% by a large margin. Code is available at \url{https://github.com/wangf3014/SCLIP}.

\end{abstract}

\section{Introduction}
\label{sec:intro}

\begin{figure}
    \centering
    \includegraphics[width=\textwidth]{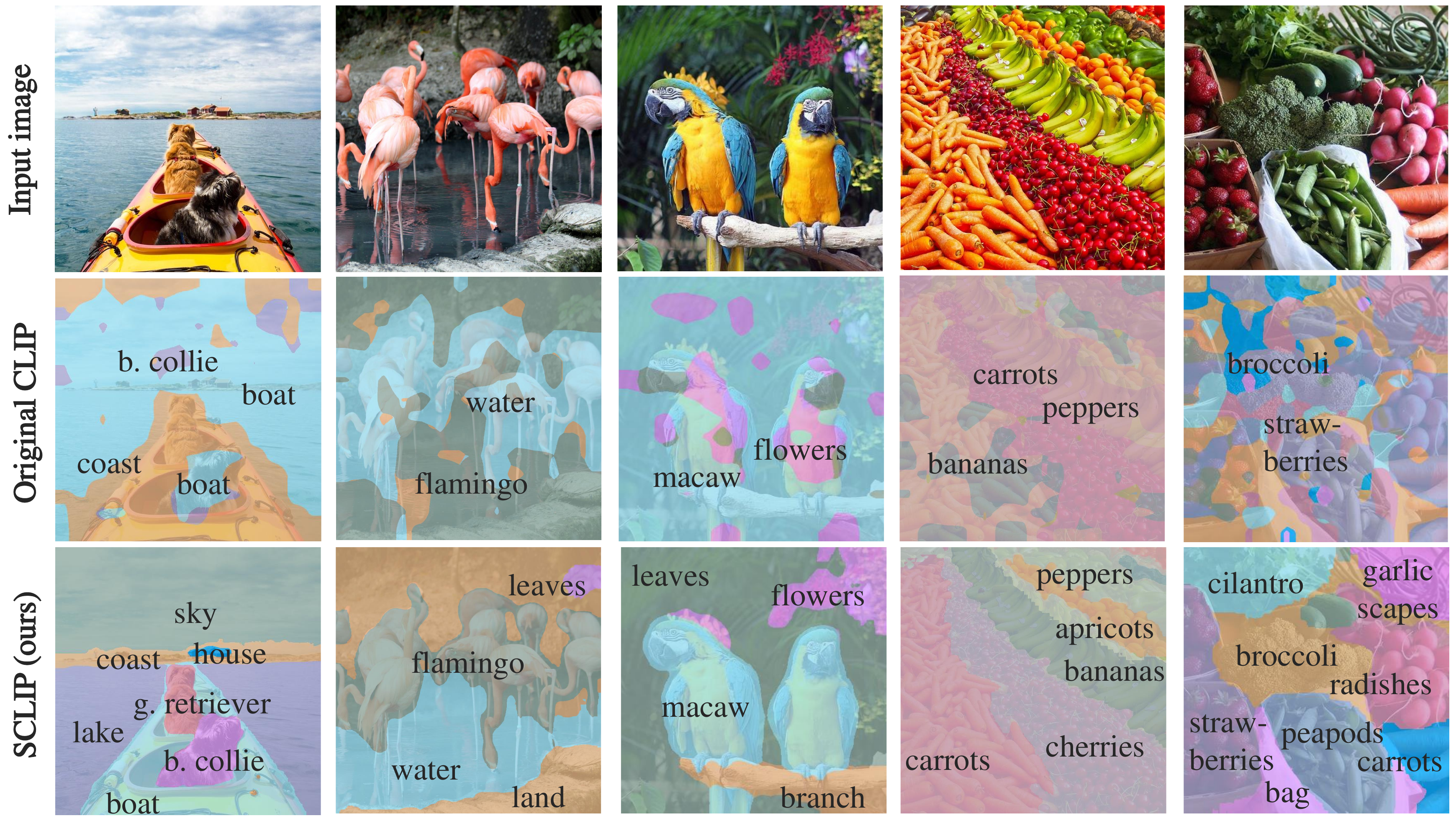}
    \caption{Open-vocabulary semantic segmentation examples. We evaluate on two images from COCO~\cite{coco} (the 3rd and the 5th examples) and three high-resolution images in the wild, where our SCLIP consistently generates high quality segmentation masks yet the original CLIP fails to correctly localize objects. We display the corresponding text query of each segmentation mask, where \textit{``g. retriever''} and \textit{``b. collie''} in the first example denote golden retriever and border collie, respectively.}
    \label{fig:samples}
\end{figure}

In the era of large foundation models, intensive pretraining followed by minimal adaptations to various downstream tasks is becoming a new paradigm for transfer learning. Nonetheless, in contrast to the significant success of foundation models in natural language processing~\cite{bert,t5,gpt3}, most visual models have yet to exhibit a comparable level of zero-shot transfer learning capabilities in various downstream tasks~\cite{fms1,fms2}. By introducing language supervision and learning on web-scale datasets, Contrastive Language-Image pretraining (CLIP) models~\cite{clip,align} are able to generalize visual representations into open-vocabulary inference and demonstrate remarkable zero-shot classification results, yet this capability remains very limited when it comes to more complex tasks such as semantic segmentation.

Specifically, CLIP performs zero-shot classification by matching image-level representations with a range of target text embeddings, by which it achieves over 70\% test accuracy on ImageNet~\cite{imagenet} when paired with proper prompting strategies~\cite{clip}. However, directly transferring this inference protocol to semantic segmentation fails to achieve favorable results. For example, when equipped with a ViT-Base/16~\cite{vit} encoder and fed with a 224$\times$224 resolution input image, CLIP can obtain a 14$\times$14 dense feature map; and by simply associating such patch-level representations with text embeddings, CLIP yields a mere 3.1\% mIoU on ADE20k~\cite{ade20k} and 5.7\% mIoU on COCO-Stuff~\cite{coco}. Considering the supervised counterparts that often produce around 40\% mIoU on this two benchmarks, this result is not really comparable. As a result, CLIP still relies on careful finetuning and in-domain adaptations for downstream dense prediction tasks~\cite{segclip,zegclip,san}.

In this work, we investigate in CLIP's potential for dense prediction and find out whether the weak supervisions of CLIP can benefit various vision tasks with minimal downstream adaptations. We start with a qualitative analysis. As shown in Figure~\ref{fig:samples}, we conduct simple open-vocabulary semantic segmentation experiments on five sample images from COCO~\cite{coco} or in the wild, where the vanilla CLIP model often presents incorrect dense predictions and noisy segmentation masks. However, despite its poor semantic segmentation performance, we find that CLIP is actually able to roughly recognize what objects appear in the image yet wrongly localizes them. For instance, in the second example, we set 10 target categories including \textit{flamingo}, \textit{water}, \textit{land}, with distractors such as \textit{sky}, \textit{building}, and \textit{person}, but although CLIP accurately obtains the correct categories such as \textit{water} and \textit{flamingo}, it predicts the opposite localizations (\ie, predicts \textit{water} for flamingos and \textit{flamingo} for water and land).

\begin{figure}[t]
    \centering
    \includegraphics[width=\textwidth]{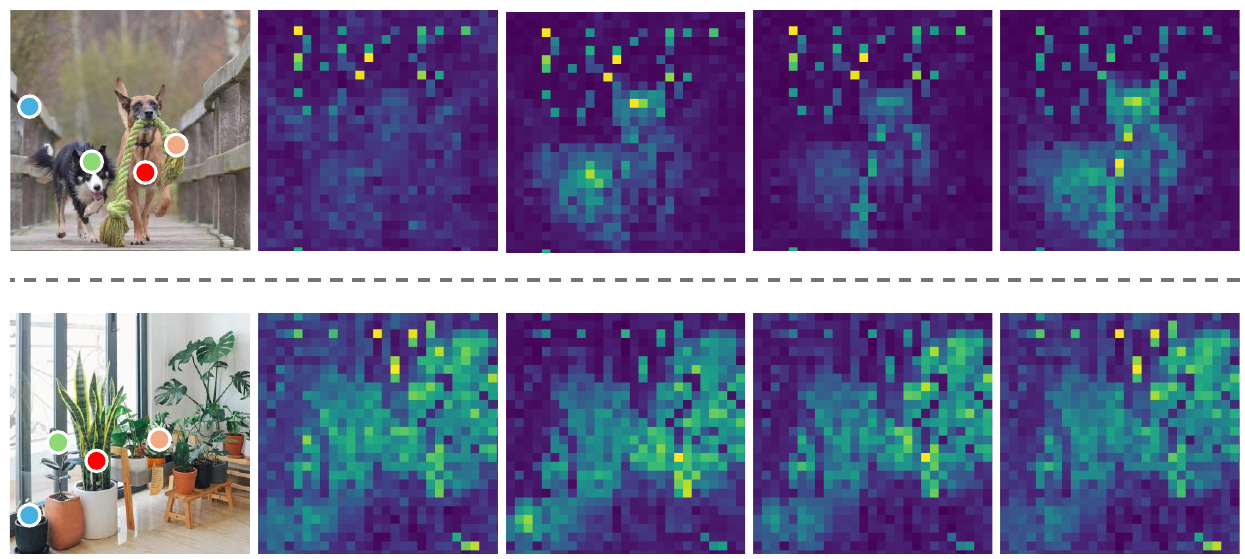}
    \caption{Final layer attention maps of vanilla CLIP with a ViT-Base/16 image encoder. We display the attention maps of four points (marked in different colors) for each example. It shows that each local visual token attends to a wide range of positions and the attention maps often share similar patterns, indicating that CLIP learns spatial-invariant visual features.}
    \label{fig:att-clip}
\end{figure}

This qualitative study suggests that the poor segmentation performance of CLIP is caused by a spatial misalignment of the patch representations, instead of a failure in extracting dense visual features. This observation makes us suspect that the problem lies in CLIP's self-attention modules because they are responsible for arranging spatial information.
In Figure~\ref{fig:att-clip}, we illustrate several examples of CLIP self-attention patterns, where each map represents the attention scores for a specific point in the image (marked in different colors). As is shown, CLIP attention maps can reflect the shape of major objects, but appears to be very similar across many different source points in the image. This suggests that CLIP learns \textbf{\textit{spatial-invariant}} visual features, implying that the local features tend to be invariant to their spatial positions in the image, and the model focuses on a holistic visual representation.

However, in dense prediction tasks like semantic segmentation, we actually desire \textbf{\textit{spatial-covariant}} features, which implies the local representations should change accordingly to their spatial positions in an image. To this end, we rethink the purpose of self-attention and introduce \textbf{C}orrelative \textbf{S}elf-\textbf{A}ttention (CSA), a novel self-attention mechanism that facilitates covariant visual features. Specifically, in contrast to the original self-attention that employs two projection matrices (\ie, the query and key) to determine attention scores, our CSA module only projects the input once to find pairwise correlations of visual tokens, which encourages each local token to attend to itself and to the positions sharing similar information with it.

Surprisingly, we find that after making this change, our CSA mechanism is very effective to adapt CLIP into dense prediction tasks. In detail, we develop our new approach \textbf{SCLIP} (\textbf{S}egmentation-adapted \textbf{CLIP} model) by employing a CSA module in place of the original self-attention block in CLIP vision encoder%
\footnote{Here we focus on transformer-based image encoders for CLIP. Compared with ResNet~\cite{resnet} encoders, the vision transformers~\cite{vit} are more suitable for zero-shot transfer into semantic segmentation, since they have 1) a global receptive field, and 2) lower down-sampling ratios (\eg, 16$\times$ for ViT-Base/16 \vs 32$\times$ for ResNet-50)}. It is noteworthy that the CSA module is not sensitive to its projection weights so we can simply reuse the pretrained parameters of original self-attention in CLIP, which makes SCLIP a tuning-free approach for semantic segmentation using a stand-alone CLIP model. 

Empirical study on our SCLIP model showcases its notable effectiveness, with yielding both impressive qualitative and quantitative outcomes: we obtain an average mIoU of 38.2\% over eight semantic segmentation benchmarks such as PASCAL Context~\cite{pascalcontext} and COCO-Stuff~\cite{coco}, substantially outperforming the existing state-of-the-art methods such as MaskCLIP~\cite{maskclip} (30.3\%), GroupViT~\cite{groupvit} (30.7\%), and TCL~\cite{tcl} (33.9\%) that support zero-shot and open-vocabulary semantic segmentation. In Figure~\ref{fig:samples}, we also show the qualitative results obtained by SCLIP for images in the COCO~\cite{coco} dataset and in the wild, where our model yields very clear and accurate segmentation masks, especially for the high-resolution inputs (\eg, the case of two dogs sitting on the boat). The primary contributions of this work can be summarized as follows: 
\begin{itemize}
    \item First, We identify the reasons of CLIP's failure in semantic segmentation, and address them by introducing a novel Correlative Self-Attention (CSA) mechanism, while extensive experiments demonstrate significant results.
    \item Next, our SCLIP approach outperforms the existing methods~\cite{maskclip,groupvit,reco,tcl} with neither fine-tuning nor any additional parameters given a pretrained CLIP model, which validates the good transferability of vision-language models in dense prediction tasks.
    \item Further, in this work, a minimal modification to CLIP yields very significant improvements in semantic segmentation, which provides with an important data point that the weakly-supervised pretraining paradigm with language guidance has very good potentials to function as a visual foundation model that supports a wide range of downstream tasks.
\end{itemize}

\section{Related Work}
\label{sec:rel}

\textbf{Transferable Visual Foundation Models.} Self-supervised pretraining has recently demonstrated good potentials in learning transferable visual representations. The models pretrained with re-constructive objectives such as masked image modeling~\cite{mae,beit,maskfeat} or discriminative objectives such as contrastive learning~\cite{moco,simclr,byol,dino,detcon,cp2} exhibit strong capabilities in adapting various visual tasks when sufficient downstream training data is available. Similarly, denoising diffusion models~\cite{diffusion,ddim,ddpm} that allow high-resolution conditional image generation and Segment Anything models~\cite{sam,seem} that facilitate semantic-agnostic image segmentation can also serve as foundation models with transferable visual features.

When incorporated with language guidance, such foundation models can be really powerful to allow open-vocabulary and zero-shot transfer learning for downstream visual tasks~\cite{clip,glide,stablediff,llava}. A representative model is CLIP~\cite{clip}, which pioneers to align visual and textual features by contrastive pretraining. Based on this, a series of follow-up works extend its scale~\cite{align,basic,florence,coca}, applications~\cite{maskclip,groupvit,vild,clipvqa}, and downstream inference protocols~\cite{coop,clipadapter,defo}.

\noindent\textbf{Open-Vocabulary Segmentation.} To fully utilize the advancements of vision-language models in zero-shot and open-vocabulary visual inference, extensive follow-up work has been initiated to investigate their applications in dense prediction tasks. For example, GroupViT~\cite{groupvit} introduces group tokens into its vision encoder and pretrains with language guidance, leading to an open-vocabulary model that well applies to semantic segmentation tasks. Also, MaskCLIP~\cite{maskclip} and CLIP Surgery~\cite{clip_surgery} make simple modification to vision transformers and enables CLIP's coarse feature localization. The study of language-guided segmentation is continuously explored~\cite{openseg,odise,segclip,zegclip,san,ovsegmentor,tcl,vilseg,reco,clippy,acseg,ifseg,loda,pacl,clipself}.

\noindent\textbf{Self-Attention for Dense Visual Features.} A series of related research has demonstrated that the potential of vision transformers in extracting dense visual features can be augmented by employing varied self-attention mechanisms. For example, in contrast to the vanilla self attention used in CLIP and conventional vision transformers~\cite{clip,vit}, the Local Attention mechanism constrains the spatial feature aggregation within a local window so that to encourage fine-grained features~\cite{swin,swinv2,biformer,crossformer++,maxvit}. Localized visual features can also be encouraged with modified self-attention mechanisms such as MaskCLIP~\cite{maskclip} which discards the processing of query and key vectors in its last transformer layer (equivalent to local attention with window being one) and MSSA~\cite{crate} that reduces the attention projections into a single matrix. In addition, some segmentation or detection-oriented transformer models leverage cross attention to map local visual features into semantic tokens~\cite{groupvit,maskformer,mask2former,detr,segmenter}. Also, the models equipped with Axial Attention~\cite{axialdeeplab,maxdeeplab,cswin} or Deformable Attention~\cite{deformabledetr,dat} demonstrate strong capabilities in dense prediction.

\section{Method}
\label{sec:method}

The central concept of our method is transforming the spatial-invariant visual features learned from the CLIP paradigm into covariant representations by architectural modifications, so that the CLIP models can generalize to dense prediction tasks. As we discussed in Section~\ref{sec:intro}, the \textbf{\textit{spatial-invariant}} features indicate that the model produces similar representations for different locations within an image and they tend to share holistic information (see Figure~\ref{fig:att-clip}), which is favorable in image-level tasks such as classification. In contrast, the \textbf{\textit{spatial-covariant}} features encourage each local token to effectively represent the visual information of its corresponding position, which is conductive to pixel-level dense prediction tasks such as semantic segmentation. We develop our approach SCLIP (Segmentation-adapted CLIP model) by introducing a new self-attention mechanism as it can re-organize the spatial information. The details can be found below.

\begin{figure}
    \centering
    \includegraphics[width=0.6\textwidth]{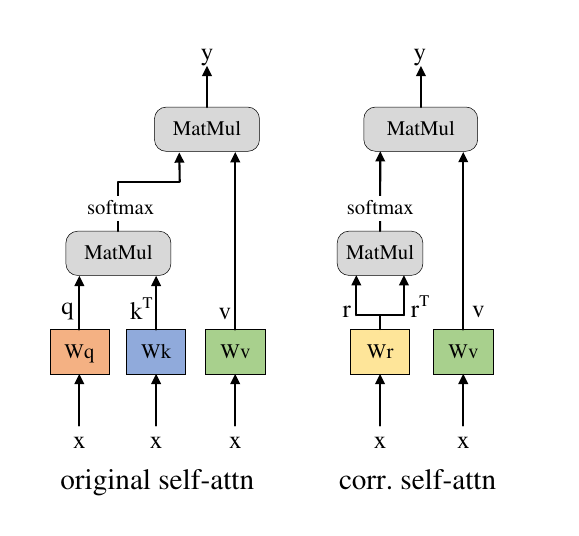}
    \caption{An architectural comparison between the original self-attention and our correlative self-attention mechanism. Our method determines attention scores by pairwise correlations between the local tokens.}
    \label{fig:corr}
\end{figure}

\subsection{Re-Visiting the Original Self-Attention}

In conventional vision transformers~\cite{vit}, each input image of size $3\times w\times h$ is initially divided into a number of non-overlapping patches, with each patch subsequently being projected into a vectorized feature $\vx_i\in\sR^d$, where $d$ denotes the dimension of the model's feature space. Each layer of the vision transformer receives a collection of visual tokens $\mX=\{ \vx_\text{cls},\vx_1, \vx_2, \dots, \vx_l \}\in\sR^{(l+1)\times d}$ as input, with $\vx_\text{cls}\in\sR^d$ denoting the class token, $l=wh/p^2$ denoting the total number of image patches ($p\times p$ size for each), and each local visual token $\vx_i\in\sR^d$ $(i=1,2,\dots,n)$ associated with a distinct position within the input image.

We illustrate the pipeline of the traditional self-attention block in Figure~\ref{fig:corr} (left). Formally, the attention map $\mA\vt\vt\vn\in\sR^{(l+1)\times(l+1)}$ is computed by
\begin{equation}
    \mA\vt\vt\vn = \text{Softmax}\left(\mX\mW_q\mW_k^T\mX^T/\sqrt{d}\right), 
\end{equation}
where $\mW_q,\mW_k\in\sR^{d\times d}$ are projection parameters learned from pretraining. Note that here we only consider the single-head self-attention for easy description. In CLIP, the vision encoders are pretrained to represent each input image by a single feature vector, which encourages the self-attention blocks to extract holistic visual representations and consequently facilitates spatial-invariant features. As mentioned above, these invariant features prevent CLIP from performing dense prediction tasks, so necessary modifications should be made for its self-attention modules to allow semantic segmentation.

A very straightforward way to this end is forcing each visual token $\vx_i$ only attending to itself, \ie, setting the attention map $\mA\vt\vt\vn$ to an identical matrix $\mI_{(l+1)\times(l+1)}$ regardless of the input. In this way, each local visual token only receives information from its corresponding position so that visual features are well localized. In practice, MaskCLIP~\cite{maskclip} uses this attention map in CLIP vision encoder's last layer and obtains a non-trivial improvement in semantic segmentation. For example, it increases CLIP's mIoU on COCO-Stuff~\cite{coco} from 5.7\% to 16.7\%. However, as this approach strictly constrains the receptive field of local tokens, the model may easily over-focus on low-level features and thus produces noisy dense predictions~\cite{maskclip,tcl}.

\begin{figure}[t]
    \centering
    \includegraphics[width=\textwidth]{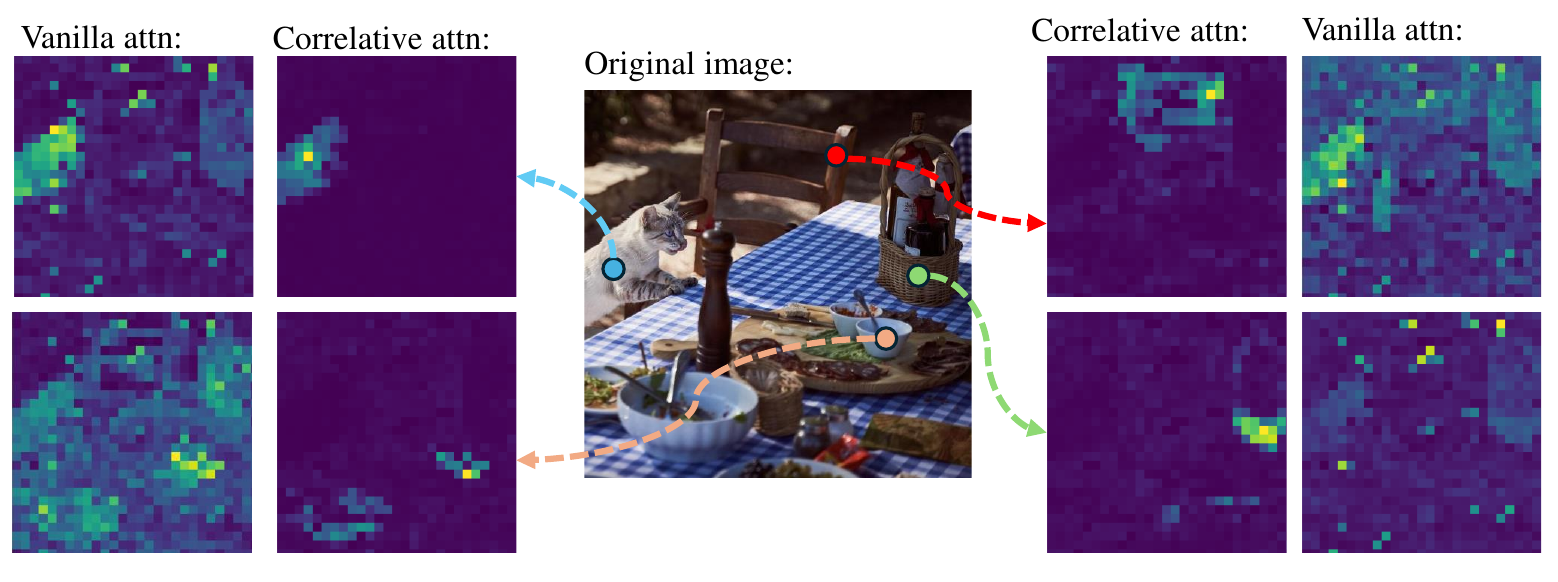}
    \caption{Comparison of attention maps. We show the attention maps of the last transformer layer in CLIP vision encoder equipped with the original self-attention (right) and our correlative self-attention (left). Our correlative self-attention exhibits spatially covariant patterns as the attention maps are distinct to different source points and show clear boundaries of semantic objects (\eg, the chair and the cat).}
    \label{fig:attn-comp}
\end{figure}

\subsection{Correlative Self-Attention}

To facilitate spatial-covariant features, we introduce \textbf{C}orrelative \textbf{S}elf-\textbf{A}ttention (CSA) mechanism which computes attention scores by pairwise correlations across local tokens, with an overall pipeline illustrated in Figure~\ref{fig:corr}. Formally, we have
\begin{equation}
    \mA\vt\vt\vn = \text{Softmax}\left(\mX\mW_r\mW_r^T\mX^T/\tau\right),
\end{equation}
where $\mX\in\sR^{(l+1)\times d}$ denotes the input and $\mW_r\in\sR^{d\times d}$ is the newly introduced projection matrix. The temperature coefficient $\tau$ is by default set to $\sqrt{d}$ following traditional self-attention. This change makes self-attention depend on the distance between the feature vectors at different positions, with an underlying idea that the tokens $\vx_i$ and $\vx_j$ assign high attention scores to each other if they have high cosine similarity after a projection. Compared with the conventional mechanism, this correlative self-attention is more suitable for dense prediction tasks for the following reasons.

First, in vision transformers, the feature localization can be intuitively reflected by the magnitude of the diagonal elements of the matrix $\mA\vt\vt\vn\in\sR^{(l+1)\times(l+1)}$. Specifically, each element $a_{ij}\in[0,1]$ of $\mA\vt\vt\vn$ measures the attention score of $\vx_i$ to $\vx_j$, so high diagonal values indicate that each local token mainly attends to its own position and the visual information of each position is consequently well localized. This explains why MaskCLIP~\cite{maskclip} works, where it forces $a_{ij}=0, i\neq j$ and $a_{ij}=1, i=j$. In the CSA module, the diagonal attention scores are also enhanced since the correlation between $\vx_i\mW_r$ and $\vx_j\mW_r$ always reaches its maximum when $i=j$ (supposing both vectors are normalized).

In addition to its notable feature localization abilities, the CSA module also thoroughly accounts for the semantic correlations across local tokens, so that it produces robust and smooth dense prediction results. Intuitively, for each local token $\vx_i$, CSA imparts high attention scores not only to $\vx_i$ itself but also to tokens that share similar semantic content. We visualize this effect in Figure~\ref{fig:attn-comp}, where for each source point, only the positions with high semantic similarity to it are assigned with noticeable attentions, and therefore the corresponding object (\eg, the chair and the cat) of each source point can be clearly recognized in the attention maps.

Further, the matrix $\mW_r$ in CSA functions as a distance measure between the features in different positions, so our model is not sensitive to the parameters of this projection layer since changing $\mW_r$ only alters the form of the distance measure. In the experiments, we find that it is unnecessary to specifically train this projection matrix, but instead manually assigning it or even employing an ensemble of randomly initialized matrices can consistently obtain very competitive results (see Section~\ref{sec:abl} and Table~\ref{tab:proj} for details). Notably, CSA's insensitivity to model parameters provides with good potentials of zero-shot adaptation into dense prediction tasks when given a pretrained CLIP model. With this merit, we can develop our segmentation model using CSA without introducing any additional parameters nor any downstream fine-tuning.

\subsection{Segmentation-Adapted CLIP Model}
\label{sec:mtd_sclip}

To develop our SCLIP approach, we employ a pretrained CLIP model with a ViT-Base/16~\cite{vit} image encoder as backbone. Generally, when it comes to adapting CLIP into a downstream task without introducing additional parameters, we actually regard its last or last several layers as a task-specific decoder head. Based on our observation of the self-attention patterns in different layers, we regard the last transformer block of CLIP's image encoder as the decoding layer to implement the adaptations while leaving the remaining components unchanged.

In this decoding layer, we replace the original self-attention block by our CSA module and reuse its parameters of $\mW_q$ and $\mW_k$ as our projection matrices. Formally, we have
\begin{equation}
\label{eq:qkr}
\begin{aligned}
    \mA\vt\vt\vn &= \text{Softmax}\left(\mX\mW_q\mW_q^T\mX^T/\tau\right) \\
    &+ \text{Softmax}\left(\mX\mW_k\mW_k^T\mX^T/\tau\right),
\end{aligned} 
\end{equation}
which makes a training-free adaptation since the matrices $\mW_q$ and $\mW_k$ can be directly loaded from CLIP.

\textbf{Post-processing of dense visual features.} In dense prediction tasks, we generally have a simple but very essential pre-hypothesis of spatial continuity, which suggests that in an image, the adjacent pixels or patches tend to share similar visual features. This prior knowledge can be easily introduced in fully supervised training as the labels of segmentation masks actually satisfy this hypothesis. However, in CLIP-like weakly supervised pretraining, there is no such explicit constraint to limit the spatial continuity of dense visual features, with only positional embeddings added in the input layer. Therefore, the existing zero-shot segmentation models often rely on specific post-processing strategies to refine or smooth their segmentation masks (\eg, PAMR~\cite{pamr} for TCL~\cite{tcl} and DenseCRF~\cite{densecrf} for ReCo~\cite{reco}).

However, we argue that such post-processing approaches should not be employed by default since ensuring the spatial continuity of output is also an integral part of the inference capability of semantic segmentation models. In our experiments, we find SCLIP to be very robust in this aspect, which does not rely on any refinement or smoothing strategies to produce good segmentation results.  

\section{Experiments}
\label{sec:exp}

\begin{table}[t]
    \centering
    \caption{Evaluation results (mIoU, \%) of our method and the baseline models on eight semantic segmentation benchmarks. The methods with an asterisk symbol \textbf{*} denote using a PAMR~\cite{pamr} post-processing strategy which introduces heavy computation cost so we \demph{de-emphasize} these results. Our results are marked in \colorbox{gray!10}{gray}. The best results on each dataset are \textbf{bolded}.}
    
    \begin{tabular}{lccccccccc}
    \toprule
    \multirow{2}{*}{\textbf{Method}} & \multicolumn{3}{c}{\it With a background category} & \multicolumn{5}{c}{\it Without background category} & \multirow{2}{*}{Avg.} \\\cmidrule(lr){2-4}\cmidrule(lr){5-9}
    & VOC21 & Context60 & Object & VOC20 & City. & Ctx59 & ADE20k & Stuff. \\\midrule
    CLIP~\cite{clip} & 18.8 & 9.9 & 8.1 & 49.4 & 6.5 & 11.1 & 3.1 & 5.7 & 14.1 \\\midrule
    MaskCLIP~\cite{maskclip} & 43.4 & 23.2 & 20.6 & 74.9 & 24.9 & 26.4 & 11.9 & 16.7 & 30.3 \\
    GroupViT~\cite{groupvit} & 52.3 & 18.7 & 27.5 & 79.7 & 18.5 & 23.4 & 10.4 & 15.3 & 30.7 \\
    ReCo~\cite{reco} & 25.1 & 19.9 & 15.7 & 57.7 & 21.6 & 22.3 & 11.2 & 14.8 & 23.5 \\
    TCL~\cite{tcl}           & 51.2 & 24.3 & 30.4 & 77.5 & 23.5 & 30.3 & 14.9 & 19.6 & 33.9 \\
    CLIP-Surg~\cite{clip_surgery} & - & - & - & - & 31.4 & 29.3 & - & 21.9 & -\\
    OVSeg.~\cite{ovsegmentor} & 53.8 & 20.4 & 25.1 & - & - & - & 5.6 & - & -\\
    SegCLIP~\cite{segclip}   & 52.6 & 24.7 & 26.5 & -    & -    & -    & -    & -    & -    \\

    \rowcolor{gray!10}
    SCLIP (ours) & \bf 59.1 & \bf 30.4 & \bf 30.5 & \bf 80.4 & \bf 32.2  & \bf 34.2 & \bf 16.1 & \bf 22.4 & \bf 38.2\\\midrule
    \multicolumn{10}{l}{\it \demph{Approaches with pamr post-processing:}}\\
    \demph{CLIP$^*$} & \demph{19.8} & \demph{8.7} & \demph{10.4} & \demph{54.2} & \demph{7.0} & \demph{11.7} & \demph{3.6} & \demph{5.9} & \demph{15.2} \\
    \demph{MaskCLIP$^*$} & \demph{52.0} & \demph{28.2}  & \demph{22.6} & \demph{72.1} & \demph{30.1} & \demph{31.5} & \demph{14.0}  & \demph{20.0} & \demph{33.8} \\
    \demph{GroupViT$^*$} & \demph{52.7} & \demph{19.5} & \demph{27.9} & \demph{81.5} & \demph{21.7} & \demph{24.4} & \demph{11.8} & \demph{16.9} & \demph{32.1} \\
    \demph{ReCo$^*$} & \demph{27.2} & \demph{21.9} & \demph{17.3} & \demph{62.4} & \demph{23.2} & \demph{24.7} & \demph{12.4} & \demph{16.3} & \demph{25.7} \\
    \demph{TCL$^*$} & \demph{55.0} & \demph{30.4} & \demph{31.6} & \demph{83.2} & \demph{24.3} & \demph{33.9} & \demph{17.1} & \demph{22.4} & \demph{37.2} \\\rowcolor{gray!10}
    SCLIP$^*$ (ours) & \bf 61.7 & \bf 31.5 & \bf 32.1 & \bf 83.5 & \bf 34.1  & \bf 36.1 & \bf 17.8 & \bf 23.9 & \bf 40.1\\
    \bottomrule
    
    \end{tabular}
    \label{tab:main}
\end{table}

\subsection{Experiment Settings}
\label{sec:expset}

\textbf{Datasets.} We evaluate our method on six commonly used semantic segmentation benchmarks, including PASCAL VOC 2012~\cite{voc12}, PASCAL Context~\cite{pascalcontext}, Cityscapes~\cite{cityscapes}, ADE20k~\cite{ade20k}, COCO-Stuff and COCO-Object~\cite{coco}. Considering the background category, we additionally evaluate on two variant datasets for PASCAL VOC and PASCAL Context. For clear reference, we denote VOC21, Context60 as the original datasets with a background class, and VOC20, Context59 as the variant without this category.
In prior works such as GroupViT~\cite{groupvit} and TCL~\cite{tcl}, they evaluate with input images resized to have a shorter side of 448 and then performing slide inference with a 448$\times$448 window and 224 stride. However, in our experiments, we find that using a smaller input size with denser sliding stride can lead to slightly higher results (\eg, 0.2\% mIoU on PASCAL Context). Specifically, we resize input images with a short side of 336 and perform slide inference with a 224$\times$224 window and 112 stride. This protocol introduces a similar level of computation as that of GroupViT, yet better fits the original input size of CLIP (\eg, 224 for ViT-Base) and is also friendly to parallel computing. For Cityscapes~\cite{cityscapes}, we resize with a 560 shorter side due to the particular high resolution of its original images. A detailed comparison of image pre-processing protocols can be found in Table~\ref{tab:pre}.

\textbf{Baselines.} CLIP~\cite{clip} is a direct baseline for our method to compare the difference of dense prediction performance between the original self-attention and our CSA mechanisms. In detail, we first extract textual embeddings of the target class names from CLIP's language encoder and then directly align them with CLIP vision encoder's dense features. We also consider the open-vocabulary semantic segmentation models derived from CLIP or similar vision language models as stronger baselines, which includes MaskCLIP~\cite{maskclip}, ReCo~\cite{reco}, and TCL~\cite{tcl}. For These methods, we report the higher numbers between our re-implementation based on their official code bases and results from the existed work~\cite{tcl}. We additionally compare them with recent baselines such as SegCLIP~\cite{segclip} and OVSegmentor~\cite{ovsegmentor}, for which we directly take the results in their original papers.

Following TCL~\cite{tcl}, we do not permit the post-processing strategies with very heavy computation cost such as Dense CRF~\cite{densecrf}, and do not consider the baselines that borrow well-pretrained models other than CLIP~\cite{odise,ovdiff}. We by default discard the Pixel-Adaptive Mask Refinement (PAMR)~\cite{pamr} technique for post-processing of segmentation masks as it also introduces intensive computation and may easily obscure the inherent inference capabilities of the segmentation models.

\subsection{Main Results}
\label{sec:main}

Table~\ref{tab:main} summarizes the comparison of various zero-shot semantic segmentation models, where our SCLIP consistently achieves the best performance across eight evaluated benchmarks, with notable leads in PASCAL Context (34.2\%), Cityscapes (32.2\%), and ADE20k (16.1\%). Overall, the average performance of SCLIP stands at 38.0\%, which is significantly higher than the second-best average performance by TCL at 33.9\%. This suggests that SCLIP provides a robust improvement over existing methods and testifies the significant effectiveness of the newly introduced correlative self-attention. Aside from the competitive baseline methods, we also report the evaluation results of the vanilla CLIP model with its original self-attention in the image encoder. As a result, this straightforward protocol fails to obtain a comparable performance as other baseline methods, indicating the incompatibility of directly transferring the original self-attention to dense prediction tasks.

In Table~\ref{tab:main} there are also results of additionally employing a PAMR post-processing layer, where almost all approaches can benefit from it with a similar level of improvements. For example, our SCLIP attains a 1.9\% average mIoU increase over the eight datasets while the baselines of GroupViT and TCL get 1.4\% and 3.3\%, respectively. Contrary to what is reported in the TCL paper~\cite{tcl} where MaskCLIP experiences a degradation in predictive performance after using the PAMR module, we find that by simply searching for suitable PAMR hyper-parameters, it can achieve a 3.5\% increase in mIoU compared with its original version. We suggest to disable this refinement strategy in the default settings of open-vocabulary segmentation since it is computationally intensive, but instead turn to some lightweight smoothing methods for the predictions.

\subsection{Ablation Study}
\label{sec:abl}

\textbf{Projection matrices in correlative self-attention.} We want to find out the effect of choosing different types of projection matrices in our correlative self-attention block. As previously discussed, the CSA module theoretically accepts any non-zero projections as its $\mW_r$, and we by default ensemble the $\mW_q$ and $\mW_k$ in CLIP's original self-attention for that (shown in Equation~\ref{eq:qkr}). Here we compare four more variants to testify its robustness.
\begin{enumerate}
    \item \textit{Identity Projection}: we directly measure the pairwise correlations by inputs $\mX$, leading to a very simple protocol of $\mA\vt\vt\vn=\text{Softmax}(\mX\mX^T/\tau)$. Note that this is not equivalent to MaskCLIP which directly forces the $\mA\vt\vt\vn$ to be an identity matrix.
    \item \textit{Ensemble of Random Initializations}: we randomly initialize several projection matrices as $\mW_r$ and then average their corresponding attention scores. Formally, we have $\mA\vt\vt\vn=\frac1n\sum_{i=1}^n\text{softmax}(\mX\mW_i\mW_i^T\mX^T/\tau)$.
    \item \textit{Projection with Single $\mW_q$ or $\mW_k$}: We load single $\mW_q$ or $\mW_k$ as our $\mW_r$ to ablate the effect of combining both.
    \item \textit{Learned Projection}: To fully exploit the potential of CSA, we specifically learn a projection matrix from the training split of each dataset. The model is able to converge well with few training samples (we use 64 for each dataset) due to the few learnable parameters.
\end{enumerate}

\begin{table}[ht]
    \centering
    \caption{Ablation results (mIoU, \%) of projection matrices in correlative self-attention. $n$ denotes the number of random projection matrices used in this experiment. Our default setting is marked in \colorbox{gray!10}{gray}. The best result on each dataset is \textbf{bolded}.}
    
    \begin{tabular}{lccc}
    \toprule
    \textbf{Mode} & PASCAL VOC & PASCAL Context & COCO-Stuff\\\midrule
    \multicolumn{4}{l}{\it Single projection matrix for CSA:}\\
    Identity projection &  57.5 & 33.0 & 21.5 \\
    $W_q$ projection & 58.2 & 33.5 & 21.7 \\
    $W_k$ projection & 58.4 & 33.1 & 21.8 \\
    Learned projection & \bf 60.4 & \bf 34.7 & \bf 22.6 \\\midrule
    \multicolumn{4}{l}{\it Random projection matrices (average of 5 trials):}\\
    $n=1$ & 57.1 & 32.4 & 20.6\\
    $n=4$ & 58.0 & 32.7 & 20.9\\
    $n=16$ & 58.1 & 32.7 & 21.2\\\midrule\rowcolor{gray!10}
    Default & 59.1 & 34.2 & 22.4\\
    \bottomrule
    
    \end{tabular}
    \label{tab:proj}
\end{table}

The results are summarized in Table~\ref{tab:proj}. Overall, the performance differences among the various modes on the three datasets are minimal, showcasing the robustness of the proposed CSA mechanism. This is particularly notable in scenarios where only a single matrix is randomly initialized, which still yields respectable results, such as an mIoU of 57.1\% on the PASCAL VOC dataset. Also, while the learned projection mode achieves the highest performance, the margin of improvement over the default training-free structure is not substantial. Given this modest gain, investing significant effort into in-domain training for learned projections may not be recommended. Excluding the learned projection, our default method consistently attains the best results. This suggests that the proposed CSA is highly compatible with the pretrained projection parameters from CLIP. This compatibility is a testament to the efficacy of CSA when paired with the robust features provided by CLIP's pretrained projections.

\textbf{Alternative approaches for feature localization.} There also exist some potential approaches to enable CLIP localizing visual features. For example, we can sharpen the attention map by simply adjusting the temperature parameters of the CLIP vision encoder, which prevents it from overly attending to global information and instead concentrates the features on a few specific positions. We denote this approach as Attention Sharpening and compare its effect to our method. Similarly, by employing local attention techniques, \ie, calculating attention scores only within a given window, we can facilitate the CLIP model in anchoring visual features to their corresponding locations. However, this method comes at the cost of losing the global receptive field inherent in the vision transformer models, preventing the model from reasoning with the assistance of tokens outside the local domain. It's noteworthy that the MaskCLIP~\cite{maskclip} algorithm can be considered as a special case of local attention when the window size is set to one. We also observe that actually the early stages of the vision transformer attend to relatively small local regions. Therefore, a possible way for CLIP feature localization is to directly borrow the attention maps in early stages in place of those in the decoding layer.

\begin{table}[t]
    \centering
    \caption{Ablation results of potential approaches for feature localization. Our default setting is marked in \colorbox{gray!10}{gray}. The best results are \textbf{bolded}.}

    \begin{minipage}{0.5\linewidth}

    \centering
    
    \begin{tabular}{lccc}
    \toprule
    \textbf{Approach} & VOC21 & Ctx59 & Stuff\\\midrule
    \multicolumn{4}{l}{\textit{Attention sharpening}}\\
    $\tau=8$ (CLIP default) & 18.8 & \bf 13.3 & \bf 5.7\\
    $\tau=2$ & \bf 21.7 & 9.5 & 4.1\\
    $\tau=0.5$ & 15.6 & 6.0 & 4.1\\
    $\tau\rightarrow 0$ (hard max) &  14.8 & 5.7 & 4.2 \\    
    \midrule

    \multicolumn{4}{l}{\it Local attention}\\
    window size $=3$ & \bf 42.9 & \bf 25.5 & \bf 16.0\\
    window size $=5$ & 30.5 & 18.3 & 8.2\\\midrule
    
    \end{tabular}
    \end{minipage}%
    \begin{minipage}{0.5\linewidth}

    \centering
    
    \begin{tabular}{lccc}
    \toprule
    \textbf{Approach} & VOC21 & Ctx59 & Stuff\\\midrule

    window size $=7$ & 28.1 & 17.9 & 8.0\\\midrule

    \multicolumn{4}{l}{\it Attention map from early stages}\\
    from layer \#1 & 41.5 & 26.2 & \bf 16.8\\
    from layer \#3 & \bf 43.0 & 26.4 & 16.2\\
    from layer \#5 & 41.7 & \bf 26.8 & 15.4\\
    from layer \#7 & 21.9 & 17.3 & 10.1\\
    \midrule\rowcolor{gray!10}

    SCLIP (ours) & \bf 59.1 & \bf 34.2 & \bf 22.4\\
    \bottomrule
    
    \end{tabular}
    \end{minipage}
    
    \label{tab:loc}
\end{table}

As summarized in Table~\ref{tab:loc}, the three alternative strategies may offer considerable enhancements over the baseline CLIP model when specific parameters are adjusted, yet they fall notably short when compared with our method. Specifically, the attention sharpening approach fails to obtain performance improvements in most cases, and only achieves a 2.9\% mIoU gain on PASCAL VOC with $\tau=2$. When we apply local attention with a window size of three, the evaluation performance is promising and almost parallels that of MaskCLIP across three different datasets. In addition, the heuristic approach of directly borrowing attention maps from early stages shows relatively better results, with a 26.8\% mIoU on PASCAL Context59 and a 16.8\% mIoU on COCO-Stuff, which even outperforms MaskCLIP.

This ablation study suggests that the mere focus on local visual features does not effectively convert a weakly supervised pre-trained model such as CLIP for semantic segmentation challenges. In contrast, our method, which incorporates a correlative self-attention mechanism that considers the relationship between local features and overarching semantic contexts, proves to be more adept for visual reasoning tasks across diverse scales.

\textbf{Image pre-processing.} As discussed in Section~\ref{sec:expset}, we adopts a new pre-processing protocol that resizes each input image with the shorter side fixed to 336 instead of 448, and perform slide inference with a smaller window of 224 and stride of 112 than previous methods~\cite{groupvit,tcl}. To ablate the effect of this protocol, we present a detailed comparison of different pre-processing strategies in Table~\ref{tab:pre}.

\begin{table}[t]
    \centering
    \caption{Ablation results of image pre-processing on PASCAL VOC. ``Img size'' in this table denotes the length of the shorter side of the resized image. Our default setting is marked in \colorbox{gray!10}{gray}. The best result is \textbf{bolded}.}
    
    \begin{tabular}{cccccccc}
    \toprule
    \textbf{Mode} & ~Img size~ & ~Window~ & ~Stride~ & ~Flops~ & ~VOC21~ & Context59 & COCO-Stuff\\\midrule
    \#1 & 224 & 224 & 112 & $1\times$ & 56.5 & 32.0 & 20.5 \\\rowcolor{gray!10}
    \#2 & 336 & 224 & 112 & $\sim4\times$ & 59.1 & 34.2 & 22.4\\
    \#3 & 336 & 336 & 112 & $\sim3\times$ & 58.6 & 34.1 & 21.9\\
    \#4 & 448 & 224 & 112 & $\sim9\times$ & \bf 60.4 & \bf 35.3 & \bf 23.4\\
    \#5 & 448 & 448 & 224 & $\sim4\times$ & 58.9 & 34.3 & 22.1\\
    \bottomrule
    
    \end{tabular}
    \label{tab:pre}
\end{table}

As is shown, in general, larger image sizes combined with smaller windows and strides lead to better performance, although they come with an increased computational cost as indicated by the higher number of Flops. Specifically, utilizing too small an image size results in substantial information loss and a marked decrease in performance, as seen with mode \#1, which achieves only a 56.5\% mIoU. Larger image sizes can enhance prediction accuracy (as demonstrated by mode \#4), yet the improvement is not significant compared to the default setting (mode \#2).

Compared to the established default settings of existing work (mode \#5), the proposed protocol achieves better results with an equivalent amount of computation. This is possibly attributed to two factors: first, CLIP inherently performs better with its original input size of 224$\times$224 pixels without fine-tuning; and second, our setting reduces the window stride, leading to smoother outputs. Furthermore, even when using the same pre-processing approach (mode \#5), our SCLIP outperforms the existing (SoTA) model, with a 58.9\% mIoU compared to TCL's 51.2\% on PASCAL VOC.

\section{Conclusion}
\label{sec:conclusion}

In this work, we propose to enhance CLIP's potentials for dense prediction tasks by introducing a novel correlative self-attention mechanism, which functions as a task-specific decoder head for semantic segmentation in our approach. The adaptation significantly improves its performance in dense vision-language inference, achieving a 38.2\% average zero-shot mIoU across eight benchmarks evaluated in this paper, outperforming the existing state-of-the-art models by a large margin. We demonstrate that minimal modifications to the existing CLIP model can yield substantial improvements in its functionality. The significant increase in zero-shot mIoU scores across various benchmarks testifies to the effectiveness of our approach. Notably, our model outperforms the existing baseline methods without any fine-tuning or additional parameters involved, which underscores the robust potential of the CLIP-like weakly-supervised pretraining paradigm in creating versatile visual foundation models.

\section*{Acknowledgements}
This work was supported by ONR N00014-23-1-2641.

%
%
\bibliographystyle{splncs04}
\bibliography{main}

\clearpage
\section*{Appendix: Additional Visualization Results}
\label{sec:apdx_vis}

\begin{figure}
    \centering
    \includegraphics[width=\textwidth]{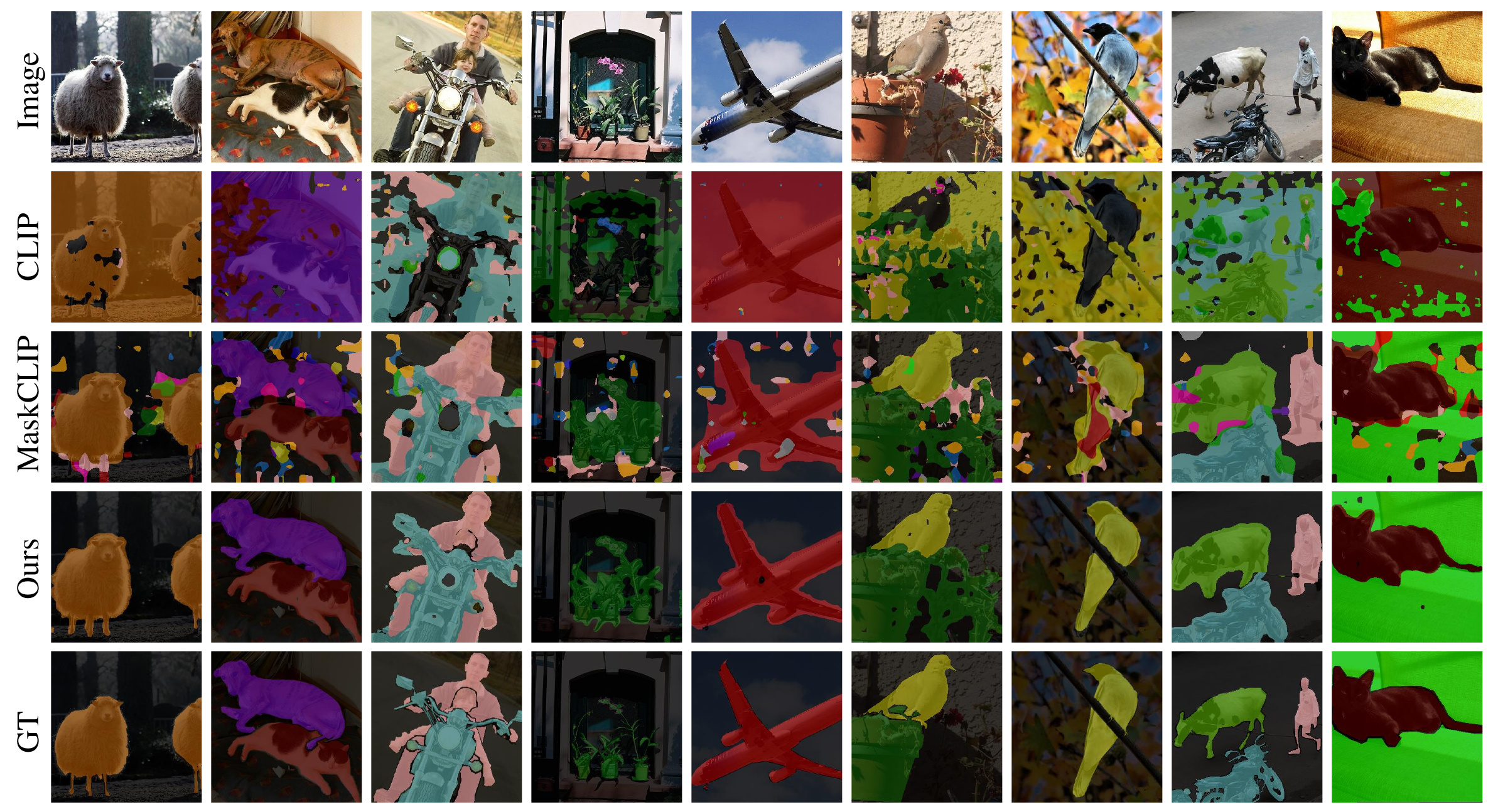}
    \caption{Additional visualization results on PASCAL VOC. ``GT'' denotes ground truth.}
    \label{fig:vis_voc}
\end{figure}

\begin{figure}
    \centering
    \includegraphics[width=\textwidth]{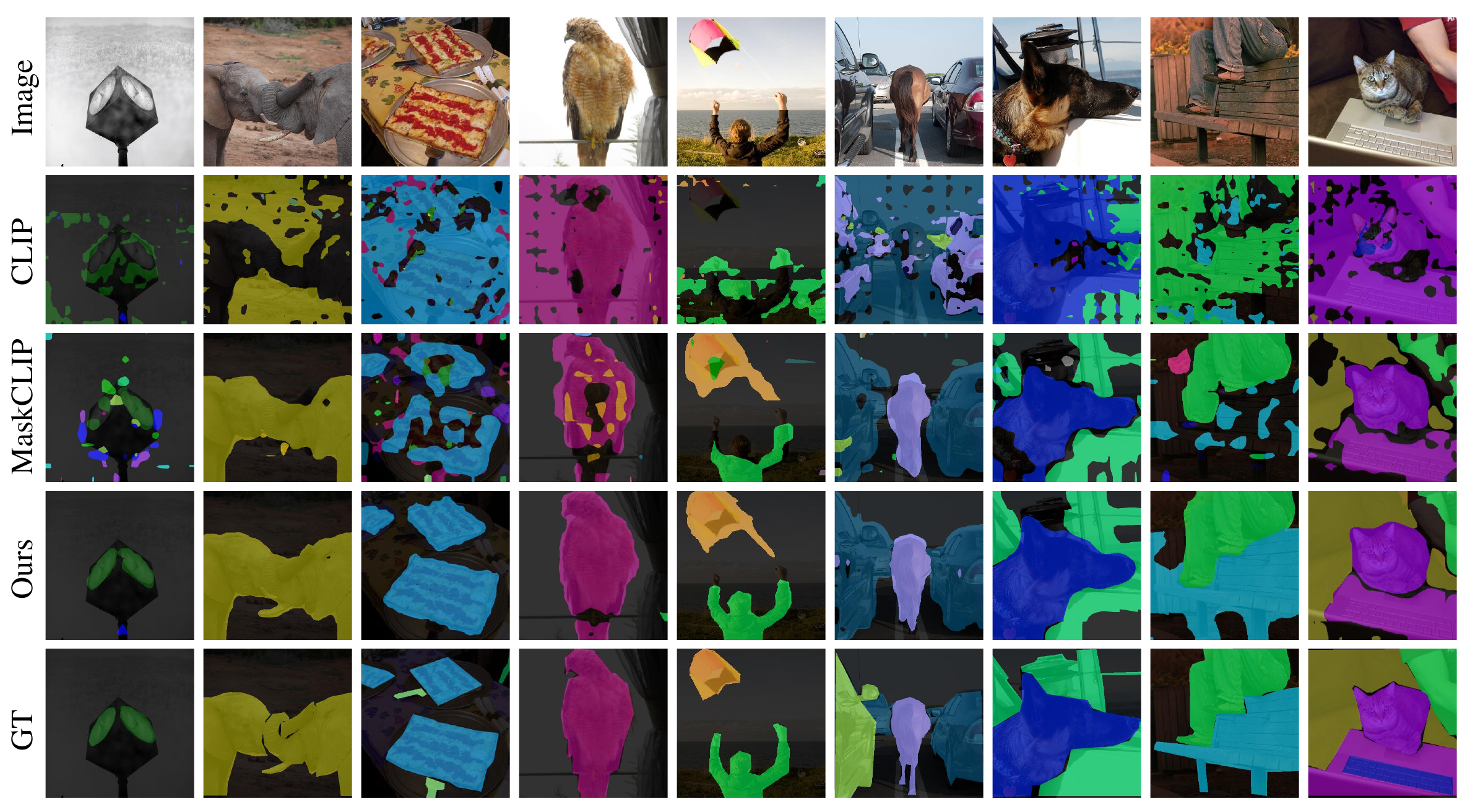}
    \caption{Additional visualization results on COCO-Object. ``GT'' denotes ground truth.}
    \label{fig:vis_obj}
\end{figure}

Here we present more qualitative results on PASCAL VOC~\cite{voc12} (Figure~\ref{fig:vis_voc}) and COCO-Object~\cite{coco} (Figure~\ref{fig:vis_obj}), and compare our model with vanilla CLIP~\cite{clip} and MaskCLIP~\cite{maskclip}. As is shown, while our SCLIP model yields very clear segmentation masks in most cases, the vanilla CLIP model fails to correctly localize the primary objects within the images, and MaskCLIP often predicts with noticeable noise and many incoherent segments.

Specifically, in datasets with relatively fewer categories such as PASCAL VOC (see Figure~\ref{fig:vis_voc}), SCLIP is able to detect very detailed semantic features. For instance, in the first example, our model accurately segments the legs of the sheep although they occupy only a very small area in the image; and in the fourth example, our segmentation mask clearly displays the shape of the branches in the potted plants, which is slightly coarser than the ground truth but significantly outperforms the result of MaskCLIP, which categorizes the area around the potted plants along with the background as a single class. These observations testify the remarkable effectiveness of our CSA module.

Semantic segmentation with more categories (\eg, 81 for COCO-Object) might be very challenging for zero-shot models. As is shown in Figure~\ref{fig:vis_obj}, in the absence of considering semantic correlations between the patch-level visual tokens, many noisy predictions emerge in MaskCLIP's segmentation results (\eg, the first and the third examples). Notably, this issue cannot be simply addressed by leveraging additional refinement or thresholding strategies, since it may lead the model to segment the image into one or very few categories and thereby degrades its inference capabilities of detailed visual features. In addition, there are several interesting observations such as in the fourth example, our segmentation mask of the bird skips the fence it stands on while the ground truth does not; and in the sixth example, the SUV on the left is annotated as ``bus'' while our model categories it to ``car''.

\end{document}